\documentclass[letterpaper, 10 pt, conference]{ieeeconf}
\IEEEoverridecommandlockouts    
\usepackage{graphics}           
\usepackage{times}              
\usepackage{amsmath}            
\usepackage{amssymb}            
\usepackage{graphicx}
\usepackage{algorithm}
\usepackage[noend]{algpseudocode}
\usepackage{booktabs}

\usepackage[font=small]{caption}

\def\secref#1{Sec.~\ref{#1}}
\def\figref#1{Fig.~\ref{#1}}
\def\tabref#1{Tab.~\ref{#1}}
\def\eqref#1{Eq.~(\ref{#1})}

\makeatletter
\usepackage{xspace}
\DeclareRobustCommand\onedot{\futurelet\@let@token\@onedot}
\def\@onedot{\ifx\@let@token.\else.\null\fi\xspace}
 
\def\ie{i.e\onedot} 
\def\cf{cf\onedot} 
 
\def\wrt{w.r.t\onedot} 
\def\etal{{et al}\onedot}
\makeatother

\usepackage{array}
\newcolumntype{L}[1]{>{\raggedright\let\newline\\\arraybackslash\hspace{0pt}}m{#1}}
\newcolumntype{C}[1]{>{\centering\let\newline\\\arraybackslash\hspace{0pt}}m{#1}}
\newcolumntype{R}[1]{>{\raggedleft\let\newline\\\arraybackslash\hspace{0pt}}m{#1}}

\newcommand{\RR}{\mathbb{R}}

\usepackage{subfigure}
\usepackage{multirow}
\usepackage{gensymb}  
\usepackage{hyperref}

\hypersetup{
    colorlinks=true,
    linkcolor=blue,
    filecolor=magenta,      
    urlcolor=blue,
}

\newcommand\Tstrut{\rule{0pt}{2.6ex}}         
\newcommand{\set}[1]{\mathcal{#1}} 	

\usepackage{color}
\definecolor{red}{rgb}{1,0,0}

\renewcommand{\v}[1]{\mathbf{#1}}	

\title{\LARGE \bf Range Image-based LiDAR Localization for Autonomous Vehicles}

\author{Xieyuanli Chen \and Ignacio Vizzo \and Thomas L\"abe   \and Jens Behley  \and Cyrill Stachniss\\
  \thanks{All authors are with the University of Bonn, Germany.}%
  \thanks{This work has partially been funded by the Deutsche Forschungsgemeinschaft (DFG, German Research Foundation) under Germany's Excellence Strategy, EXC-2070 - 390732324 - PhenoRob, 
  and by the Chinese Scholarship Committee.
  }%
}

\begin{document}
\maketitle
\thispagestyle{empty}
\pagestyle{empty}

\begin{abstract}
  Robust and accurate, map-based localization is crucial for autonomous mobile systems.
  In this paper, we exploit range images generated from 3D LiDAR scans to address the problem of localizing mobile robots or autonomous cars in a map of a large-scale outdoor environment represented by a triangular mesh.
  We use the Poisson surface reconstruction to generate the mesh-based map representation.
  Based on the range images generated from the current LiDAR scan and the synthetic rendered views from the mesh-based map, we propose a new observation model and integrate it into a Monte Carlo localization framework, which achieves better localization performance and generalizes well to different environments.  
  We test the proposed localization approach on multiple datasets collected in different environments with different LiDAR scanners.
  The experimental results show that our method can reliably and accurately localize a mobile system in different environments and operate online at the LiDAR sensor frame rate to track the vehicle pose.
\end{abstract}

\section{Introduction}
\label{sec:intro}

Precise localization is a fundamental capability required by most autonomous mobile systems. With a localization system, a mobile robot or an autonomous car is capable to estimate its pose in a map based on observations obtained with onboard sensors.
Precise and reliable LiDAR-based global localization is needed for autonomous driving, especially in GPS-denied environments or situations where GPS cannot provide accurate localization results.

Most autonomous mobile systems have a 3D LiDAR sensor onboard to perceive the environment and directly provide 3D range measurements.
In this paper, we tackle the problem of vehicle localization based on such 3D LiDAR sensors.
For localization, probabilistic state estimation techniques are used in most localization systems today. 
In particular, particle filters are a versatile tool as they do not need to restrict the motion or observation model to follow a specific distribution, such as a Gaussian. 
When utilizing particle filters, we need to design an appropriate observation model in lieu with a map representation. 
Frequently used observation models for LiDARs are the beam-end point model, also called the likelihood field~\cite{thrun2005probrobbook}, the ray-casting model~\cite{dellaert1999icra}, or models based on handcrafted features~\cite{steder2011iros, zhang2018iros}.
These methods either only work efficiently with 2D LiDAR scanners or need carefully designed features to work properly.
Recently, researchers also focused on data-driven learning of such observation models~\cite{barsan2018corl, wei2019cvpr, chen2020iros}, which provide accurate and reliable results as long as the environment is close to the environment used for learning the model.

\begin{figure}[t]
  \vspace{0.2cm}
  \centering
    \includegraphics[width=0.99\linewidth]{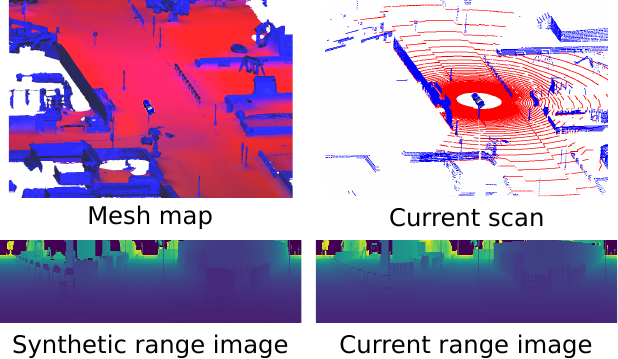}
  \caption{Visualization of range images and triangular mesh map. On the left, we show the triangular mesh used as the map and the rendered synthetic range image from the mesh. In the mesh map, red parts correspond to ground planes and blue parts represent non-ground structures. On the right, we show the LiDAR point cloud at the same location and the corresponding range image generated from the LiDAR scan.} 
  \label{fig:motivation}
\end{figure}

Instead of using raw point clouds obtained from a 3D LiDAR sensor or features generated or learned from the point clouds, we investigate range images for 3D LiDAR-based localization for autonomous vehicles.
As shown in~\figref{fig:motivation}, we project the point clouds into range images and localize the autonomous system with rendered views from a map that is represented with a triangular mesh.
There are several reasons to use range image representation and maps represented by a mesh. The cylindrical range image is a natural and light-weight representation of the scan from a rotating 3D LiDAR, and a mesh map is more compact than a large point cloud. Those properties enable our approach to achieve global localization in large-scale environments. Furthermore, the rendering of range images from a mesh map can be performed efficiently using computer graphics techniques.
Therefore, range images and mesh maps are a perfect match for achieving LiDAR-based global localization.

The main contribution of this paper is a novel observation model for 3D LiDAR-based localization. Our model is based on range images generated from both, the real LiDAR scans and synthetic renderings of the {\emph{mesh}} map. We use the difference between them to formulate the observation model for a Monte Carlo localization~(MCL) for updating the importance weights of the particles. Based on our novel observation model, our approach provides $(x, y, \theta)$-pose estimates for the vehicle and achieves global localization using 3D LiDAR scans. Furthermore, our approach generalizes well over different environments collected with different types of LiDAR scanners.
In sum, we make three key claims:
Our approach is able to
(i) achieve global localization accurately and reliably using 3D LiDAR data,
(ii) can be used for different types of LiDAR sensors, and, 
(iii) generalize well over different environments.
These claims are backed up by the paper and our experimental evaluation.
{The source code of our approach is available at:
\url{https://github.com/PRBonn/range-mcl}}.

\section{Related Work}
\label{sec:related}

For localization given a map, one often distinguishes between pose tracking and global localization.
In pose tracking, the vehicle starts from a known pose and the pose is updated over time. 
In global localization, no pose prior is available. 
In this work, we address global localization using 3D LiDAR data without assuming any pose prior from GPS or other sensors.
Therefore, we concentrate here mainly on LiDAR-based approaches.

In the context of autonomous cars, many approaches were proposed for accurate pose tracking using multiple sensor modalities and high-definition~(HD) maps. 
Levinson \etal~\cite{levinson2007rss} utilize GPS, IMU, and LiDAR scans to build HD maps for localization. 
They generate a 2D surface image of ground reflectivity in the infrared spectrum and define an observation model that uses these intensities. 
The uncertainty in intensity values has been handled by building a prior map~\cite{wolcott2015icra}.  
Barsan \etal~\cite{barsan2018corl} use a fully convolutional neural network~(CNN) together with HD maps to perform online-to-map matching for improving the robustness to dynamic objects and eliminating the need for LiDAR intensity calibration. 
Merfels and Stachniss~\cite{merfels2016iros} present an efficient chain-like pose graph for vehicle localization exploiting graph optimization techniques and different sensing modalities. 
Based on this work, Wilbers \etal~\cite{wilbers2019icra} propose a LiDAR-based localization system performing a combination of local data association between laser scans and HD map features, temporal data association smoothing, and a map matching approach for robustification.
The approaches above show good performance for tracking vehicles' poses but require GPS information for operating on HD maps. In contrast, our approach addresses global localization using only 3D LiDAR data without assuming any pose prior.

To achieve global localization, traditional approaches rely on probabilistic state estimation techniques~\cite{thrun2005probrobbook}. 
A popular framework is Monte Carlo localization~\cite{dellaert1999icra,thrun2001ai,fox1999aaai}, which uses a particle filter to estimate the robot's pose and is widely used in robot localization systems~\cite{bennewitz2006euros,chen2020iros,kuemmerle2014jfr,sun2020icra,yan2019ecmr}.

Recently, several approaches exploiting deep neural networks and semantic information for 3D LiDAR localization have been proposed. For example, 
Ma \etal~\cite{ma2019iros} combine semantic information such as lanes and traffic signs in a Bayesian filtering framework to achieve accurate and robust localization within sparse HD maps.  
Yan \etal~\cite{yan2019ecmr} exploit buildings and intersections information from a LiDAR-based semantic segmentation system~\cite{milioto2019iros} to localize in OpenStreetMap data. Schaefer \etal~\cite{schaefer2019ecmr} detect and extract pole landmarks from 3D LiDAR scans for long-term urban vehicle localization whereas
Tinchev \etal~\cite{tinchev2019ral} propose a learning-based method to match segments of trees and localize in both urban and natural environments.
Sun \etal~\cite{sun2020icra} use a deep-probabilistic model to accelerate the initialization of the Monte Carlo localization and achieve a fast localization in outdoor environments.
In our previous work~\cite{chen2020iros, chen2020rss}, we also exploit CNNs with semantics to predict the overlap between LiDAR scans as well as their yaw angle offset, and use this information to build a learning-based observation model for Monte Carlo localization.
The learning-based methods perform well in the trained environments, while they usually cannot generalize well in different environments or different LiDAR sensors.

In sum, our method only uses LiDAR data to achieve global localization outdoors without using any GPS. Moreover, our approach uses range information directly without exploiting neural networks, semantics, or extracting landmarks. 
Therefore, it generalizes well to different environments and different LiDAR sensors and does not require new training data when moving to different environments.

\section{Our Approach}
\label{sec:main}

\begin{figure}[t]
  \centering
  \vspace{0.2cm}
    \includegraphics[width=0.99\linewidth]{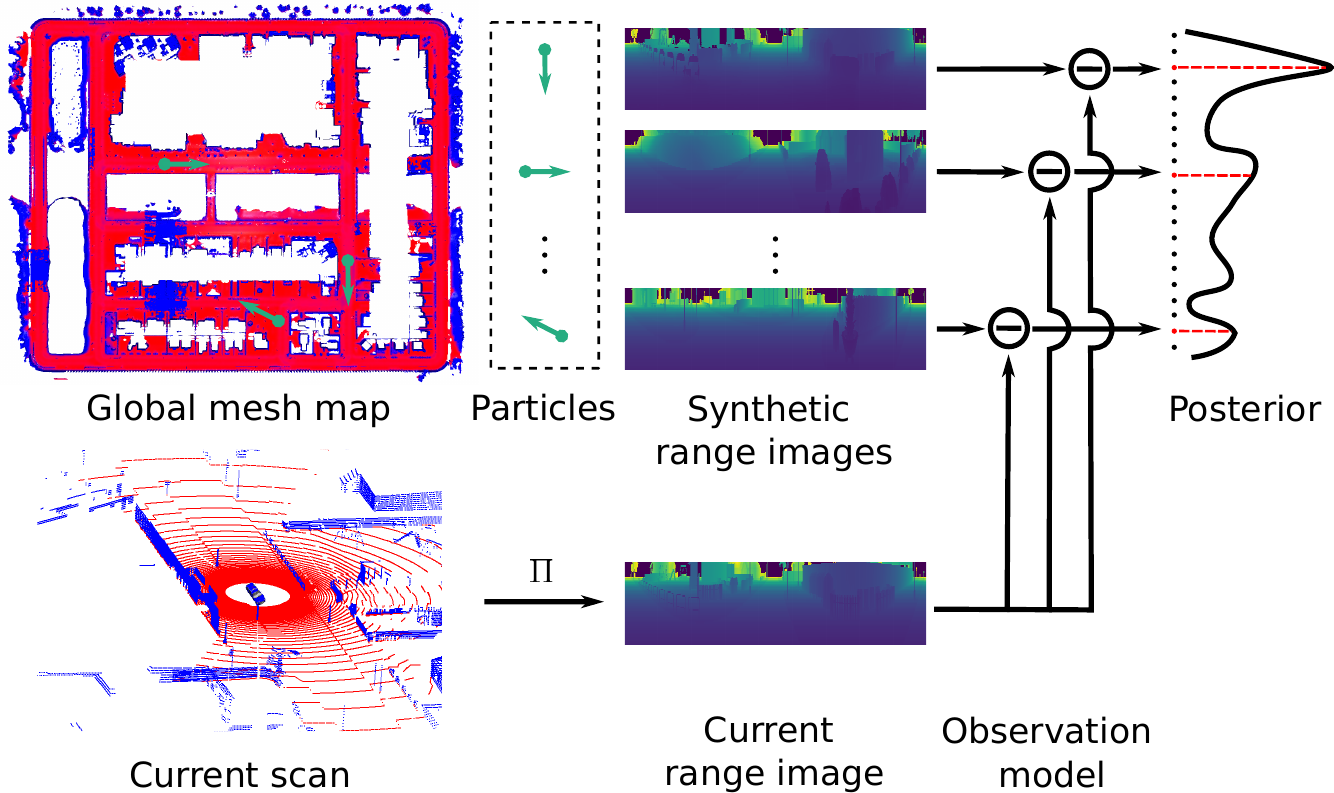}
  \caption{Overview of our approach. We project the LiDAR point cloud into a range image and compare it to synthetic range images rendered at each particle location from a mesh map. Based on the range images, we propose a new observation model for localization and integrate it into a Monte Carlo localization system to estimate the pose posterior of the vehicle.}
  \label{fig:overview}
\end{figure}

In this paper, we propose a probabilistic global localization system for autonomous vehicles using a 3D LiDAR sensor, see \figref{fig:overview} for an illustration.
To this end, we project the LiDAR point cloud into a range image~(see \secref{sec:range image generating}) and compare it to synthetic range images rendered at each particle location from a map represented by a triangular mesh~(see \secref{sec:approach-mapping} and \secref{sec:rendering}).
Based on the range images, we propose a new observation model for LiDAR-based localization~(see \secref{sec:observation-model}) and integrate it into a Monte Carlo localization system~(see \secref{sec:approach-mcl}).
Furthermore, we employ a tile map to accelerate the rendering and decide when the system converges~(see \secref{sec:approach-tile}).
Using an OpenGL-based rendering pipeline, the proposed system operates online at the frame rate of the LiDAR sensor after convergence.

\subsection{Range Image Generation}
\label{sec:range image generating}

The key idea of the proposed method is to use range images generated from LiDAR scans and rendered from the triangular mesh map for robot localization.
To generate range images, we use a spherical projection~\cite{behley2018rss, milioto2019iros, chen2019iros, chen2020rss}.
We project the point cloud $\set{P}$ to the so-called vertex map $\set{V}_D: \RR^2 \mapsto \RR^3$, where each pixel contains the nearest 3D point.
Each point $\v{p}_i = (x, y, z)$ is converted via the function $\Pi:  \RR^3 \mapsto \RR^2$ to spherical coordinates and finally to image coordinates $(u,v)$, \ie,
\begin{align}
  \left( \begin{array}{c} u \vspace{0.0em}\\ v \end{array}\right) &= \left(\begin{array}{cc} \frac{1}{2}\left[1-\arctan(y, x) \cdot \pi^{-1}\right] \cdot w   \vspace{0.5em}\\  
  \left[1 - \left(\arcsin(z\cdot r^{-1}) + \mathrm{f}_{\mathrm{up}}\right) \mathrm{f}^{-1}\right] \cdot h \end{array} \right), \label{eq:projection}
\end{align}
where $r = ||\v{p}||_2$ is the range, $\mathrm{f} = \mathrm{f}_{\mathrm{up}} + \mathrm{f}_{\mathrm{down}}$ is the vertical field-of-view of the sensor, and $w, h$ are the width and height of the resulting vertex map $\set{V}_D$.
Given the vertex map $\set{V}_D$ and range $r$ of points at each coordinate $(u,v)$, we generate the corresponding range image $\set{R}_D$,
on which the subsequent computations are based upon.

\subsection{Mesh-based Map Representation}
\label{sec:approach-mapping}
We use a triangular mesh as a map $\set{M}$ of the environment.
A triangular mesh provides us with a compact representation that enables us to render the aforementioned range images at the frame rate of the LiDAR sensor.

To generate the map, we use LiDAR scans together with their poses provided a SLAM system uses for mapping. Note that it is not necessary for our approach to use the same LiDAR sensor for map generation and localization.
We employ Poisson surface reconstruction~(PSR)~\cite{kazhdan2006eg} to obtain the representation of the map as a triangular mesh from point clouds.
The~PSR algorithm requires an oriented point cloud, thus normals for all the points in the input cloud. To estimate the surface normals, we use a range image-based normal estimation~\cite{behley2018rss}, which computes normals in the so-called normal map $\set{N}_D$ for each coordinate $(u,v)$  using cross products over forward differences of the corresponding vertex map pixel, \ie,
\begin{align}
  \set{N}_D((u,v)) &= \left(\set{V}_D((u + 1, v)) - \set{V}_D((u, v))\right) \nonumber\\
		   &  \quad \times \left(\set{V}_D((u, v + 1)) - \set{V}_D((u, v))\right).
\end{align}

To decrease the storage size of the map, we use a ground segmentation algorithm.
We first compute the empirical covariance matrix of all the points in the cloud. We then compute the normalized Eigenvectors of~$\Sigma:\v{e}_1, \v{e}_2, \v{e}_3$ with corresponding Eigenvalues $\lambda_1 \ge \lambda_2 \ge \lambda_3$. 
We use a simple, yet effective, approach to label a point $\v{p}_i = (x, y, z)$ with corresponding normal $\v{n}_i$ as ground. $\v{p}_i$ is considered to belong to the ground surface if it satisfies the following criteria: $\v{n}_i \cdot \v{e}_3 > \cos{\alpha_{\text{thres}}}$ and $z < z_{\text{thres}}$.

After all ground points have been labeled, we proceed to run the reconstruction algorithm that retains the labels encoded as two different colors of the vertices $\v{p}_i$ in the reconstructed triangles. 
To simplify the ground mesh, we split the complete mesh into non-ground and ground mesh. 
For the ground mesh, we contract all vertices to a single vertex that are inside a voxel of a given size $s_{\text{voxel}}$. 
Then, we filter the ground vertices using an average filter by replacing each vertex $\v{v}_i$ with $\v{v}^*_i$ averaging all adjacent vertices \mbox{$v_n \in \mathcal{N}$}:
\begin{align}
  \v{v}^*_i &= \frac{\v{v}_i + \sum_{n \in \mathcal{N}} \v{v}_n}{|\mathcal{N}| + 1} \,.
\end{align}

After averaging, invalid edges are removed. Once the ground surface has been simplified, it is combined with the rest of the mesh with no further processing.
This simplification allows us to decrease the size of the mesh model to about~$50\%$ of its original size.

The PSR algorithm provides a global solution that considers the whole input data at once without resorting to heuristic partitioning~\cite{carr2001siggraph} or blending~\cite{alexa2001vis, shen2004siggraph}. Therefore, instead of building the map incrementally~\cite{vizzo2021icra}, we aggregate all the oriented point clouds into a global reference frame using map poses, then, this oriented point cloud map is directly fed to the PSR algorithm, yielding a globally consistent triangular mesh of the environment.

\begin{figure}[t]
  \centering
  \vspace{0.1cm}
    \includegraphics[width=0.80\linewidth]{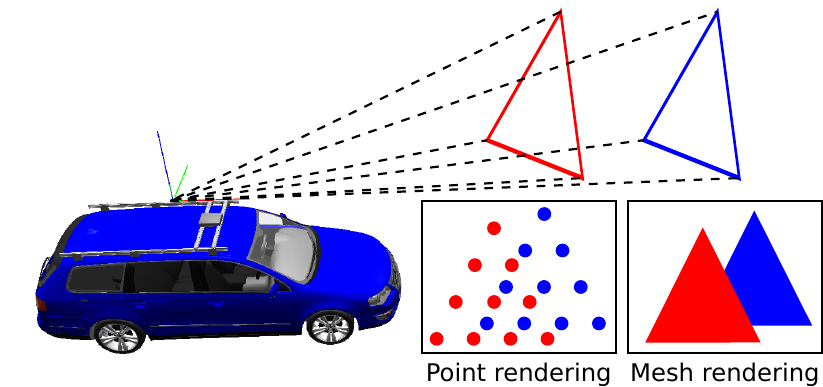}
  \caption{Rendering example. In contrast to a map represented by a point cloud, triangular meshes are smoother and more compact. During the projection, for each triangle, only three vertices need to be projected. Moreover, triangles can better represent the occlusion relationship between different objects.}
  \label{fig:rendering}
  \vspace{-0.2cm}
\end{figure}

\subsection{Rendering Synthetic Range Images}
\label{sec:rendering}
Given a particle $j$ with its state vector~$(x^j, y^j, \theta^j)$ and the triangular mesh map $\set{M}$, we use OpenGL to render a synthetic range image for that particle. 
Using the spherical projection, \cf ~\eqref{eq:projection}, we project vertices of the triangles from the given particle pose and let OpenGL shade the triangle surface considering the occlusion, as shown in \figref{fig:rendering}.
To further accelerate rendering, we render batches of range images for multiple particles using instancing of the map, which allows us to avoid reading vertex positions multiple times and minimizes the number of draw calls.

\subsection{Monte Carlo Localization}
\label{sec:approach-mcl}

Monte Carlo localization~(MCL) is commonly implemented using a particle filter~\cite{dellaert1999icra}. 
In our case, each particle represents a hypothesis for the autonomous vehicle's 2D pose $\v{x}_t = (x, y, \theta)_t$ at time~$t$. 
When the robot moves, the pose of each particle is updated based on a motion model with the control input~$\v{u}_t$. 
The expected observation from the predicted pose of each particle in the map~$\mathcal{M}$ is then compared to the actual observation~$\v{z}_t$ acquired by the robot to update the particle's weight based on an observation model. 
Particles are resampled according to their weight distribution and resampling is triggered whenever the effective number of particles drops below a specific threshold~\cite{grisetti2007tro}.
After several iterations of this procedure, the particles eventually converge around the true pose.

MCL realizes a recursive Bayesian filter estimating a probability density~$p(\v{x}_t\mid\v{z}_{1:t},\v{u}_{1:t})$ over the pose~$\v{x_t}$ given all observations~$\v{z}_{1:t}$ up to time~$t$ and motion controls~$\v{u}_{1:t}$ up to time~$t$. This posterior is updated as follows:
\begin{align}
  &p(\v{x}_t\mid\v{z}_{1:t},\v{u}_{1:t}) = \eta~p(\v{z}_t\mid\v{x}_{t}, \set{M}) \cdot\nonumber\\
  &\;\;\int{p(\v{x}_t\mid\v{u}_{t}, \v{x}_{t-1})~p(\v{x}_{t-1} \mid \v{z}_{1:t-1},\v{u}_{1:t-1})\ d\v{x}_{t-1}},
\label{eq:bayesian}
\end{align}
where~$\eta$ is a normalization constant, $p(\v{x}_t\mid\v{u}_{t}, \v{x}_{t-1})$~is the motion model, $p(\v{z}_t\mid\v{x}_{t}, \set{M})$~is the observation model, and $p(\v{x}_{t-1} \mid \v{z}_{1:t-1},\v{u}_{1:t-1})$~is the probability distribution for the prior state $\v{v}_{t-1}$.

In this work, we focus on the observation model and employ a standard motion model for vehicles~\cite{thrun2005probrobbook}.

\subsection{Range Image-based Observation Model}
\label{sec:observation-model}

\begin{figure}[t]
  \vspace{0.2cm}
  \centering
  \subfigure[Location heatmap]{\includegraphics[width=0.40\linewidth]{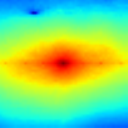}}
  \hspace{0.02\linewidth}
  \subfigure[Heading likelihood]{\includegraphics[width=0.42\linewidth, height=0.40\linewidth]{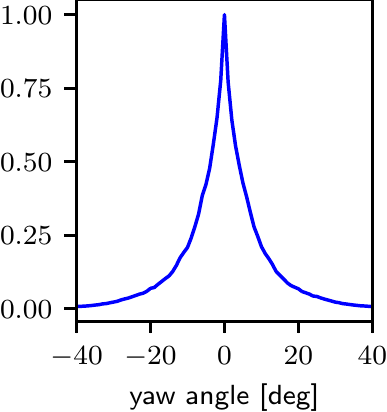}}
  \subfigure[Scene used to generate the heatmap]{\includegraphics[width=0.88\linewidth]{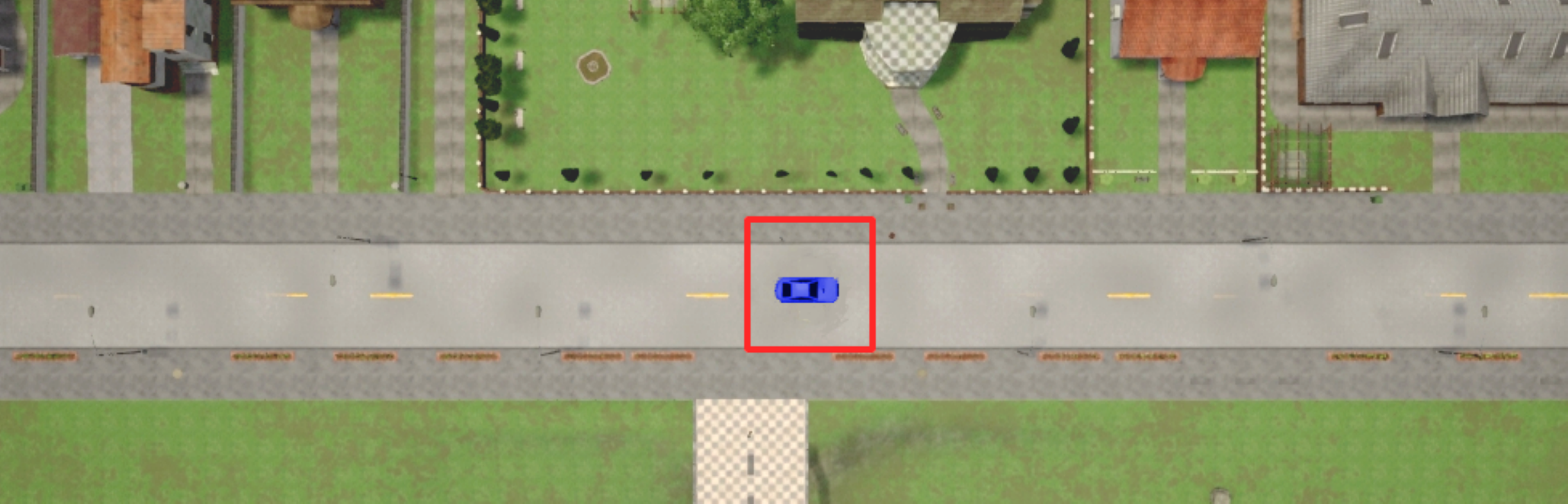}}
  \caption{Range image-based observation model. (a) A local heatmap shows the location likelihood of the scan at the car's position with respect to the map with the same heading. Red shades correspond to higher weights. (b) Heading likelihood of the observation model when changing the yaw angle with the same location. (c) A top-down view of the Carla scene used in this example.}
  \label{fig:sensor_model}
  \vspace{-0.2cm}
\end{figure}

Based on the generated range image from the current LiDAR scan and the rendered synthetic range images for all particles, we design an observation model.

Each particle~$j$ represents a pose hypothesis \mbox{$\v{x}_t^j = (x, y, \theta)_t^j$} at time~$t$. 
Given the corresponding synthetic rendered range image $\v{z}^{j}$ for the $j$-th particle rendered at the particle's pose hypothesis, we compare it to the current range image $\v{z}_{t}$ generated from the LiDAR point cloud. 
The likelihood $p\left(\v{z}_{t} \mid \v{x}_{t}, \set{M} \right)$ of the $j$-th particle is then approximated using a Gaussian distribution:
\begin{align}
  p \left(\v{z}_{t} \mid \v{x}_{t}, \set{M} \right) &\propto \exp{\left(-\frac{1}{2} \frac{{d\left(\v{z}_{t}, \v{z}^{j} \right)}^2}{\sigma^2_d}\right)},
\label{eq:sensor}
\end{align}
where~$d$ corresponds to the difference between or similarity of the range images $\v{z}_{t}$ and $\v{z}^{j}$.

There are several ways to calculate this similarity.
For example, we could directly compare two range images at pixel level with absolute differences or using a cross-correlation.
One could also generate features and compare two images in the feature space. 
Recently, there are also many deep learning-based algorithms proposed~\cite{barsan2018corl, chen2020iros, sun2020icra, wei2019cvpr}.

In this paper, our goal is to investigate the use of range images generated from LiDAR scans and triangular meshes for a Monte Carlo localization system and the particular choice of similarity computation is application dependent.
To keep the whole system fast and easy to use, we opted for a fast-to-compute and effective method and use $d = N^{-1}\sum |\v{z}_t - \v{z}^j|$, \ie, the mean of the absolute pixel-wise differences, where $N$ is the number of valid pixels in the current range image. 
Our results show that this choice is effective and generalizes well to datasets collected in different environments with different types of LiDAR sensors (see~\secref{sec:generalization}).

\figref{fig:sensor_model} shows the probabilities in a local area calculated by our proposed observation model and shows that it encodes the pose hypotheses very well.
In contrast to our prior work~\cite{chen2020rss}, which decouples the observation model into two parts, location likelihood and heading likelihood, the proposed observation model can estimate the likelihood for the whole state space~$\v{x}_t = (x, y, \theta)_t$ at once using one model, which is elegant and fast.

\subsection{Tiled Map Representation}
\label{sec:approach-tile}

We split the global mesh-based map into tiles to accelerate the Monte Carlo localization by more efficient rendering.
We only use parts of the mesh associated to a tile, which are close to the particle position.
Besides more efficient rendering, we also use tiles to determine when the localization has converged. 
If all particles are localized in at most $N_{\text{conv}}$ tiles, we assume that the localization has converged and reduce the number of particles to track the pose. 
Tiles also enable the runtime of our method to be independent of the size of the whole environment after converging.

\section{Experimental Evaluation}
\label{sec:exp}

We present our experiments to show the capabilities of our method and to
support our claims, that our approach is able to:
(i) achieve global localization accurately and reliably using 3D LiDAR data,
(ii) can be used for different types of LiDAR scanners, and,
(iii) generalize well to different environments without changing parameters.

\textbf{Implementation Details.} 
We implement our code based on Python and OpenGL.
To generate a triangular mesh map, we use the PSR implementation from Open3D~\cite{zhou2018arxiv}.
For ground point extraction, we use $\alpha_{\text{thres}} = 30^\circ$ and $z_{\text{thres}}$ as the sensor mounted height, and employ a voxel grid with voxel sizes $s_{\text{voxel}}$ = 1.0$\, m$.
We use tiles of size~$100\times100\,$m$^2$ and we reduce the number of particles from initially $10,000$ to~$100$ particles after convergence, \ie, only $N_{\text{conv}} = 1$ tile is covered by particles. The size of the tile map and the reduced number of particles are trade-offs between runtime and accuracy.
We set $\sigma_d = 5$ in~\eqref{eq:sensor}, and we only update the weights of particles when the car is moving.  
All parameters are tuned with one dataset (IPB-Car) and kept the same for all other experiments with different datasets and sensors. 

\textbf{Datasets.} We evaluate the generalization ability of our method using multiple datasets, including Carla~\cite{dosovitskiy2017corl}, IPB-Car~\cite{chen2020iros}, MulRan (KAIST)~\cite{kim2020icra} and Apollo (Columbia-Park)~\cite{zhou2020eccv-apollo}.
These datasets are collected in different environments with different types of LiDAR scanners over different times, see~\tabref{tab:dataset} for more details. For the Carla simulator, we added objects for the test sequences, which are not present in the map to simulate a changing environment like in the real datasets.
For all experiments on different datasets, we only change the intrinsic and extrinsic calibration parameters of the LiDAR sensors for the generation of the range images and keep all other parameters, especially those of the MCL, the same.

\begin{table}[t]
  \centering
  \vspace{0.2cm}
  \caption{Dataset Overview}
  \label{tab:dataset}
\begin{tabular}{ll|ccc}
\toprule
\multirow{2}{*}{Dataset} & \multirow{2}{*}{Sensor} & \multirow{2}{*}{Sequence} & Acquisition  & \multirow{2}{*}{Length} \\ 
 &  &  & time  & \\ \midrule
\multirow{2}{*}{\parbox{1.3cm}{Carla (Simulator)}} & \multirow{2}{*}{\parbox{1.5cm}{8 - 128 beam LiDAR}} & map (0.2\,Gb) & n/a & 3.5\,km \\
 &  & 00 & n/a & 0.7\,km \\ \midrule
\multirow{3}{*}{\parbox{1.3cm}{IPB-Car (Germany)}} & \multirow{3}{*}{Ouster 64} & map (0.8\,Gb) & 02/2020 &  6.2\,km \\
 &  & 00 & 09/2019 & 1.7\,km \\
 &  & 01 & 11/2019 & 1.9\,km \\ \midrule
\multirow{2}{*}{\parbox{1.2cm}{MulRan (Korea)}} & \multirow{2}{*}{Ouster 64} & map (0.5\,Gb) & 08/2019 & 6.0\,km \\
 &  & 00 & 06/2019 & 6.1\,km \\ \midrule
\multirow{2}{*}{\parbox{1.2cm}{{Apollo (U.S.)}}} & \multirow{2}{*}{Velodyne 64} & map (5.4\,Gb) & 09/2018 & 44.8\,km \\ 
 &  & 00 & 10/2018 & 8.8\,km \\ 
 
 \bottomrule
\end{tabular}
\end{table}

\textbf{Baselines.}
In the following experiments, we use the same MCL framework and only change the observation models. 
We compare our method with three baseline observation models: the typical beam-end model~\cite{thrun2005probrobbook}, a histogram-based model derived from the work of R\"ohling \etal~\cite{roehling2015iros}, and a deep learning-based model~\cite{chen2020iros}. 

The beam-end observation model is often used for 2D LiDAR data.
For 3D LiDAR scans, it needs many more particles to make sure that it converges to the correct pose, which causes the computation time to increase substantially. In this paper, we implement the beam-end model with a down-sampled point cloud map using voxelization with a resolution of~$10$\,cm. 

Our second baseline for comparison is inspired by R\"ohling \etal~\cite{roehling2015iros}, which exploit the use of similarity measures on histograms extracted from 3D LiDAR data. 

The third baseline is the overlap-based localization~\cite{chen2020iros}. It uses a deep neural network to estimate the overlap and yaw angle offset between a query scan and map data, and on top of this builds an observation model for MCL. The overlap-based method utilizes a grid map and stores virtual frames for each grid cell. We refer to the corresponding paper~\cite{chen2020iros} for more details.

\begin{figure}[t]
  \centering
  \vspace{0.2cm}
  \includegraphics[width=0.95\linewidth]{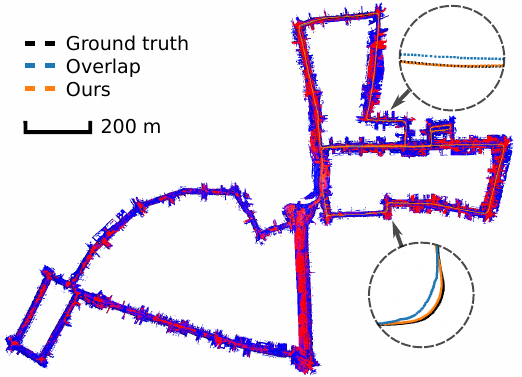}
  \caption{Localization results using $10,000$ particles on the IPB-Car dataset. Shown are the mesh map, the ground truth trajectory (black), the overlap-based result (blue), and the result of our proposed method (orange).}
  \label{fig:trajectories}
  \vspace{-0.2cm}
\end{figure}

\subsection{Localization Performance}
\label{sec:loc_results}

The experiment presented in this section investigates the localization performance of our approach. It supports the claim that our approach achieves global localization accurately and reliably using 3D LiDAR data.

For qualitative results, we show the trajectories of the localization results tested on the IPB-Car dataset in~\figref{fig:trajectories}. The results illustrate that the proposed method localizes well in the map using only LiDAR data collected in dynamic environments at different times.
Comparing to the baseline methods, the proposed method tracks the pose more accurately.

\begin{figure}[t]
  \vspace{0.2cm}
  \centering
  \hspace*{0.7cm}\includegraphics[width=0.85\linewidth]{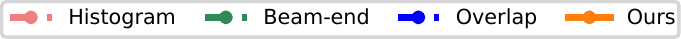}\\
  \includegraphics[width=0.48\linewidth]{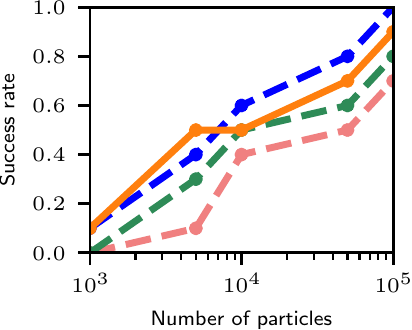}
  \includegraphics[width=0.48\linewidth]{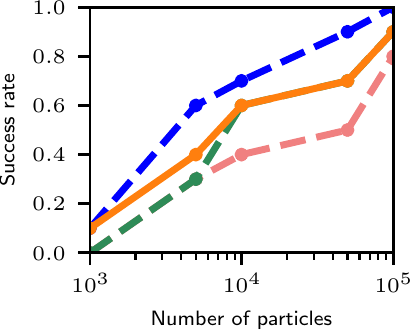}
  \caption{Success rate of the different observation models for~$10$ globalization runs. Here, we use sequence 00 (left) and sequence 01 (right) to localize in the map of the IPB-Car dataset.}
  \vspace{-0.2cm}
  \label{fig:success_rate}
\end{figure}

For quantitative results, we first calculate the success rate for different methods with different numbers of particles comparing our approach to the aforementioned methods, see~\figref{fig:success_rate}. 
We tested the methods using five different numbers of particles~$N=$~\{$1,000$, $5,000$, $10,000$, $50,000$, $100,000$\}. For each setup, we sample $10$~trajectories and perform global localization. 
The x-axis represents the number of particles, while the y-axis is the success rate of different setups. 
The success rate for a specific setup of one method is calculated using the number of success cases divided by the total number of the tests. 
To decide whether one test is successful or not, we check the location error by every $100$~frames after convergence. 
If the location error is smaller than $5\,$m, we count this run as a success.

Quantitative results of localization accuracy are shown in~\tabref{tab:loc_results}.
The upper part shows the location error of all methods tested with both sequences. 
The location error is defined as the root mean square error~(RMSE) of each test in terms of $(x,\,y)$ Euclidean error with respect to the ground truth poses. It shows the mean and the standard deviation of the error for each observation model. Note that the location error is only calculated for success cases with $10,000$ particles. 
The lower part shows the yaw angle error. It is the RMSE of each test in terms of yaw angle error with respect to the ground truth poses. The table shows the mean and the standard deviation of the error for each observation model. As before, the yaw angle error is also only calculated for cases in which the global localization converged with $10,000$ particles.

The quantitative results show that our method outperforms all baseline methods in location accuracy while achieving similar heading accuracy. 
The reason is that our method uses online rendered range images and does not rely on discrete grids. 
Therefore, our method will not be affected by the resolution of the grid.
However, this means that our method requires more particles to achieve the same success rate.
Thus, we use a large amount of particles for initialization. 
It will then achieve a high success rate while without influencing the runtime after convergence, due to the use of tiles, see in~\secref{sec:runtime}.

\begin{table}[t]
\centering
\caption{Localization results on the IPB-Car dataset}
\scalebox{0.92}{
\scriptsize{
\begin{tabular}{cccccc}
\toprule
Sequence & \multicolumn{4}{c}{Location error [m]} \\
\cline{2-5}
 & Beam-end & Histogram-based & Overlap-based & Ours \Tstrut \\
\midrule
00 & $0.92$~$\pm$~$0.27$      & $1.85$~$\pm$~$0.34$ & $0.81$~$\pm$~$0.13$ & $\mathbf{0.66 \pm 0.12}$\\
01 & $0.67$~$\pm$~$0.11$ & $1.86$~$\pm$~$0.34$ & $0.88$~$\pm$~$0.07$ & $\mathbf{0.44 \pm 0.03}$\\
\midrule
\midrule
Sequence & \multicolumn{4}{c}{Yaw angle error [deg]} \\
\cline{2-5}
 & Beam-end & Histogram-based & Overlap-based & Ours \Tstrut \\
\midrule
00 & $1.87$~$\pm$~$0.47$ & $3.10$~$\pm$~$3.07$ & $1.74$~$\pm$~$0.11$ & $\mathbf{1.69 \pm 0.11}$\\
01 & $2.10$~$\pm$~$0.59$ & $3.11$~$\pm$~$3.08$ & $\mathbf{1.88 \pm 0.09}$ & $2.53$~$\pm$~$0.79$ \\
\midrule
\end{tabular}
}
}
\label{tab:loc_results}
\vspace{-0.2cm}
\end{table}

\begin{table}[t]
\centering
\vspace{0.2cm}
\caption{{Localization results on datasets using different sensors.}}
\scalebox{0.95}{
\scriptsize{
\begin{tabular}{llcc}
\toprule
Dataset & Scanner & Location RMSE [m] & Yaw angle RMSE [deg] \\
\midrule
\multirow{4}{*}{Carla} & $8$-beams   & $0.48$ & $3.87$ \\
	  				   & $16$-beams  & $0.43$ & $3.87$ \\
	  				   & $32$-beams  & $0.42$ & $3.40$ \\
	  				   & $64$-beams  & $0.36$ & $3.46$ \\
	  				   & $128$-beams & $0.33$ & $3.31$ \\
\midrule
MulRan  & Velodyne $64$ & $0.83$ & $3.14$  \\
\midrule
Apollo  & Ouster $64$   & $0.57$ & $3.40$  \\
\midrule
\end{tabular}
}
}
\label{tab:differentLidarResults}
\vspace{-0.2cm}
\end{table}

\subsection{Generalization}
\label{sec:generalization}

The experiment supports the claim that our method is able to use different types of LiDAR sensors to localize in the same mesh map.
We test $5$ different types of LiDAR sensors in the Carla simulator, including Quanergy MQ-8 (8-beams), Velodyne Puck (16-beams), Velodyne HDL-32E (32-beams), Velodyne HDL-64E (64-beams), and Ouster OS1-128 (128-beams). We use the real parameters from the aforementioned  LiDAR sensors, which include the number of beams and the field of view.
As shown in~\tabref{tab:differentLidarResults}, our method works well with all different types of sensors and achieves good localization results even with relatively sparse scans~(location RMSE of $0.48\,$m with the 8-beam LiDAR).

\begin{figure}[t]
  \centering
  \includegraphics[width=0.75\linewidth]{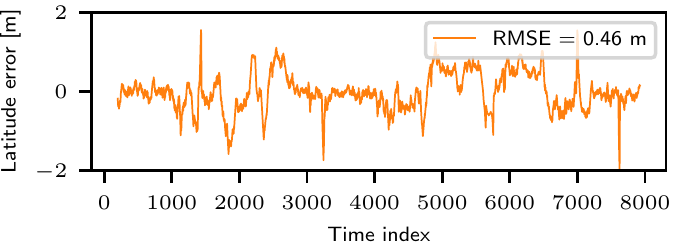}
  \includegraphics[width=0.75\linewidth]{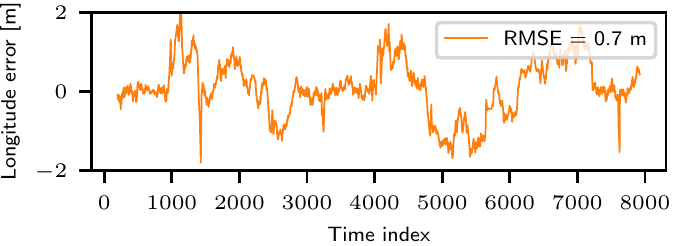}
  \includegraphics[width=0.75\linewidth]{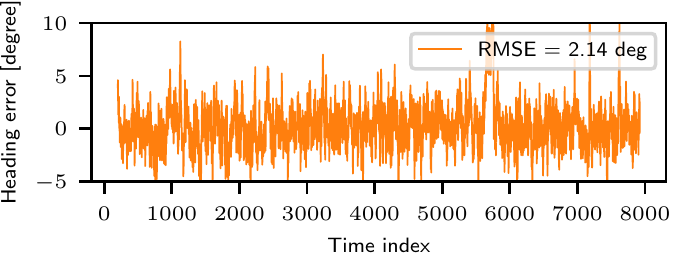}
  \caption{Localization results on the MulRan dataset~\wrt provided GPS locations. The top figure shows the latitude error, the middle figure shows the longitude error, and the bottom figure shows the heading error.}
  \label{fig:mulran_loc_results}
  \vspace{-0.2cm}
\end{figure}

\tabref{tab:differentLidarResults} and~\figref{fig:mulran_loc_results} also verify the claim that our method generalizes well over different environments.
We test our method on both the MulRan and Apollo datasets with the same parameters used in the Carla and IPB-Car dataset.
Our method works well also in Korean and U.S. urban environments.
 
\subsection{Runtime}
\label{sec:runtime}

Here we show that our 
approach runs fast enough to support online processing on the robot at sensor frame rate. 
We tested our method on a regular computer with an Intel i7-8700 with 3.2 GHz and an Nvidia GeForce GTX 1080 Ti with 11~GB of memory.
On the Carla dataset, before convergence, the maximum time for one query frame is $56.7$\,s with~$10,000$ particles. 
After convergence, the average frame rates of our method is $21.8$\,Hz with~$100$ particles and the tile maps size of~$100\times100\,$m$^2$.

\section{Conclusion}
\label{sec:conclusion}

In this paper, we presented a novel range image-based online LiDAR localization approach.  
Our method exploits range images generated from LiDAR scans and a triangular mesh-based map representation.
This allows us to localize autonomous systems in the given map successfully.
We implemented and evaluated our approach on different datasets
and provided comparisons to other existing techniques.
The experiments suggest that our method is able to achieve global localization reliably and accurately. 
Moreover, our method generalizes well to different environments and can be used with different LiDAR sensors.
After the localization convergence, our method can operate online at the sensor frame rate.

\bibliographystyle{plain}

\bibliography{glorified,new}

\end{document}